\documentclass{article}

\PassOptionsToPackage{table}{xcolor}
\PassOptionsToPackage{numbers,sort&compress}{natbib}
\usepackage[preprint]{neurips_2024}

\usepackage[utf8]{inputenc} 
\usepackage[T1]{fontenc}    
\usepackage{hyperref}       
\usepackage{url}            
\usepackage{booktabs}       
\usepackage{amsfonts}       
\usepackage{nicefrac}       
\usepackage{microtype}      
\usepackage{xcolor}         
\usepackage{graphicx}
\usepackage{array}
\usepackage{amsmath}
\usepackage{comment}

\newcommand{\myparagraph}[1]{\vspace{2pt}{\noindent\textbf{#1}}}
\newcommand{\pearson}{\operatorname{R}}

\newcommand{\cov}{\operatorname{cov}}
\newcommand{\sgn}{\operatorname{sgn}}

\newcommand{\Exp}{\operatorname{Exp}}
\newcommand{\Norm}{\mathcal{N}}
\newcommand{\p}{\operatorname{p}}

\newcommand{\x}{\mathbf{x}}
\newcommand{\y}{\mathbf{y}}
\newcommand{\z}{\mathbf{z}}
\newcommand{\G}{\mathbf{G}}
\newcommand{\tauv}{\tau}
\newcommand{\tauw}{\tau_\text{w}}
\newcommand{\taug}{\mathring{\tauw}}

\RequirePackage{xspace}
\makeatletter
\DeclareRobustCommand\onedot{\futurelet\@let@token\@onedot}
\def\@onedot{\ifx\@let@token.\else.\null\fi\xspace}

\def\eg{\emph{e.g}\onedot}

\makeatother

\title{Back to the Basics on Predicting Transfer Performance}

\author{%
  Levy Chaves\thanks{The authors contributed equally.}~$^{,1}$, Eduardo Valle\footnotemark[1]~$^{,2,3}$, Alceu Bissoto$^{1}$, Sandra Avila$^{1}$\\
  $^{1}$Recod.ai Lab, Institute of Computing, Universidade Estadual de Campinas, Brazil\\
  \texttt{\{levy.chaves, alceubissoto, sandra\}@ic.unicamp.br}\\
  $^{2}$Recod.ai Lab, School of Electrical and Computing Engineering,\\ Universidade Estadual de Campinas, Brazil\\
  $^{3}$Valeo.ai, Paris, France \\
    \texttt{eduardo.valle@valeo.com} \\
}

\begin{document}

\maketitle

\begin{abstract}
  In the evolving landscape of deep learning, selecting the best pre-trained models from a growing number of choices is a challenge. Transferability scorers propose alleviating this scenario, but their recent proliferation, ironically, poses the challenge of their own assessment. In this work, we propose both robust benchmark guidelines for transferability scorers, and a well-founded technique to combine multiple scorers, which we show consistently improves their results. We extensively evaluate 13 scorers from literature across 11 datasets, comprising generalist, fine-grained, and medical imaging datasets. We show that few scorers match the predictive performance of the simple raw metric of models on ImageNet, and that all predictors suffer on medical datasets. Our results highlight the potential of combining different information sources for reliably predicting transferability across varied domains.

\end{abstract}

\section{Introduction}
\label{sec:intro}

The success of transfer learning presents to deep-learning practitioners a challenge: selecting, among dozens of pre-trained source models, the most suitable one for their specific needs. Fine-tuning all candidates to compare them defeats one of the purposes of transfer learning: rationalizing the use of computing.
Transferability estimation is an efficient alternative, identifying promising source models through heuristics, forgoing expensive fine-tuning.

Given such advantage, it is no surprise that literature on transferability estimation is advancing fast, with an increasing offer of transferability scorers. That poses the challenge to benchmark the scorers themselves, while opening the opportunity to combine different scorers, as they exploit different aspects of transfer. 

Those are the aims of this work, whose contributions we summarize below:

\myparagraph{Benchmarking.} We propose a novel approach for the rigorous benchmarking transferability scorers by applying principles of statistical design of experiments (Section~\ref{sec:benchmarking}). By employing a bootstrapping procedure and a novel combined analysis (see below), we mitigate the issue of benchmarking under limited experimental data that affects existing literature. We also emphasize the importance of using the source model performance on the source task as a baseline.

\myparagraph{Combined Analysis.} Existing literature employs a separate analysis of scorers for each dataset, each with very few data points. We propose a novel metric that allows a combined analysis for all datasets, and, thus, more significant estimation of the scorers' capabilities (Section~\ref{sec:benchmark_standard}).

\myparagraph{Back to Bayes.} We propose a multi-level Bayesian regression as a well-founded mechanism to combine multiple scorers (Section~\ref{sec:predictive}). In addition to its theoretical appeal, that mechanism offers a good compromise between the flexibility necessary to conciliate ``conflicting opinions'' and the rigidity necessary to avoid overfitting under a very low-data regimen. 

\begin{figure*}[ht!]
    \centering
    \includegraphics[width=\textwidth]{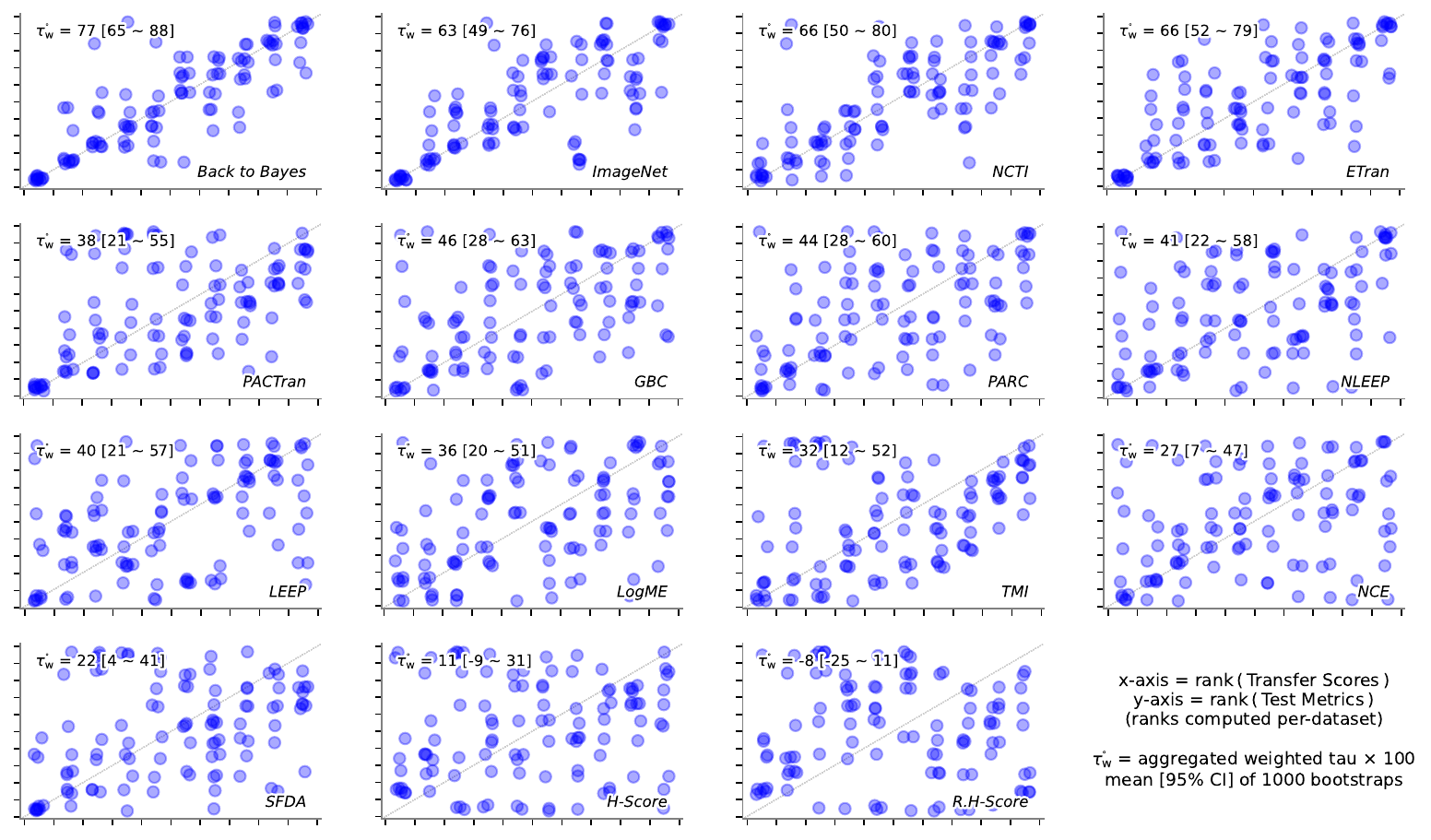}
    \caption{SOTA scorers, ImageNet-based baseline, and Back to Bayes. The plots illustrate the techniques proposed in Section~\ref{sec:benchmarking}: the combined analysis of all datasets and the use of bootstrapping. The combined analysis is possible due to the proposed aggregated tau (Section~\ref{sec:benchmark_standard}), and the bootstrapping is employed to compute its 95\%-confidence intervals, shown inside the brackets on each plot. Rather than individual data points, the main message of those plots is whether the data points concentrate at the main diagonal of the plot, showing the scorer's ability of matching the ranks of transfer scores and test metrics.}
    \label{fig:aggregated_plots}
\end{figure*}

\myparagraph{SOTA evaluation.} We apply our proposed benchmark to 13 transferability scorers found in the literature, evaluating them for 11 datasets, comprising generalist, fine-grained, and medical datasets.  The literature review appears in Section~\ref{sec:related}, and the results of our experiments in Section~\ref{sec:results}.

\section{Related Work}
\label{sec:related}

We assume the reader understands the basics of transfer learning and defer to a recent survey for the fundamentals~\cite{zhuang2020comprehensive}. That said, we must clarify that the meaning of  ``transferability'' across literature comprises at least: \textit{task transferability}, which measures the relatedness of tasks (\eg, classification, segmentation) under the same data~\cite{taskonomy, dwivedi2019representation}; \textit{dataset transferability}, which concerns the selection of the best datasets under the same model architecture~\cite{achille2019task2vec, vu2020exploring, gao2021information, fifty2021efficiently, kim2023taskweb, du2024datamap}; \textit{architecture transferability}, which concerns the choice of different model architectures under the same pre-training source dataset~\cite{leep,nce,hscore,reghscore,nleep,parc,sfda,logme,gbc,pactran,etran,ncti,tmi,transrate,tan2021otce}; and \textit{checkpoint transferability}, in which both model architecture and source dataset may vary~\cite{nleep}. Our scope will be architecture transferability. 

\myparagraph{Transferability scorers} aim to suggest the best transfer scheme. Within our scope, that means suggesting best model architecture for a given target dataset by exploiting attributes of both.

Measuring the performance of idealized classifiers is a common strategy. LEEP~\cite{leep} was an early adopter, by simulating an idealized classifier on probabilities estimated from source and target class labels. The limitations of such label-based technique led followers to prefer employing the features extracted by the old model on the new data, \eg, NLEEP~\cite{nleep}, which adapts the probabilistic model of LEEP to exploit a Gaussian mixture model learned on such features. SFDA~\cite{sfda} uses the features to learn a Fisher Discriminant Analysis, robustified with hard examples, noise augmentation, and a class-separability criterion. LogME~\cite{logme} learns a Bayesian linear classifier and uses the logarithm of its maximum evidence as the transfer score. PARC~\cite{parc}, inspired by earlier works~\cite{dwivedi2019representation, dwivedi2020duality} that used a “probe” model pre-trained on the target data, avoided the choice (and training expense) of the probe by simulating it with a fixed embedding function on the ground-truth labels.

Statistics on pre-trained data conditioned on the new class labels are another common strategy. While NCE~\cite{nce} estimates the conditional entropy between the source and target labels, using features (as explained above) is much more common. H-Score~\cite{hscore} measured the intra- \textit{vs.} inter-class variance of features, while R.H-Score~\cite{reghscore} is a shrinkage-regularized variant of the same estimator.
GBC~\cite{gbc} estimates the feature overlap using the Gaussian Bhattacharyya Coefficient. TMI~\cite{tmi} goes against the (explicit or implied) preference of the above methods for low intra-class variance, arguing that class-conditional entropy is a proxy for model adaptability.

Optimal-transport techniques~\cite{alvarez2020geometric,tan2021otce, gao2021information} move further towards a precise estimation of the similarity of source and target datasets in feature space. Those methods assess the individual features of both source and target samples, and thus, require the former, and have quadratic complexity.

PACTran~\cite{pactran} is a unique method that estimates the theoretically grounded PAC-Bayesian bound~\cite{mcallester1998some}, on three choices of priors (Gaussian, Dirichlet, and gamma) for a posterior that only accounts for the last layer of the network. Nevertheless, PACTran works well in practice for fully fine-tuned models.

NCTI~\cite{ncti} is another literature outlier, exploiting the theoretical foundation of \textit{neural collapse}~\cite{papyan2020prevalence}, which predicts that deep models reach a specific geometric configuration when fully trained. 

ETran~\cite{etran} is one of the few hybrid methods  in the literature. It uses a simulated simplified classifier based on Linear Discriminant Analysis to evaluate the compatibility of the source model features with the target dataset classes. ETran also computes an energy/entropy-inspired evaluation of the separability of the feature space. The final score is a simple sum of the two components.

Most scorers justify their design on the much simpler scenario of a frozen feature extractor and only the last-layer classifier retrained, as the theoretical underpinning of full-model fine-tuning is much more complex. Nevertheless, most scorers provide at least an empirical evaluation under full-model fine-tuning, since it almost invariably leads to better results for downstream tasks.

\myparagraph{Combined and learned scoring.} Scorers exploit different theoretical and empirical evidence from transfer schemes, suggesting that a single scorer is unlikely to succeed in all cases. However, combined schemes such as ETran~\cite{etran} are seldom found. In task transferability, the success of Taskonomy~\cite{taskonomy} and Task2Vec~\cite{achille2019task2vec} demonstrates the potential of learning for transfer prediction, but learning is conspicuously absent from architecture transferability. In this work, we address both gaps.

\myparagraph{ImageNet metric as scores.} ~\citet{kornblith2019better} found a robust correlation between the source model top-1 accuracy during its pre-training on ImageNet and the test metric on a diverse array of target datasets.~\citet{ericsson2021well} further confirmed those findings for self-supervised pre-training and few-shot and object-detection downstream tasks. Yet, LEEP~\cite{leep} is the only scorer we found that is measured against ImageNet raw scores, and even then, for only one of the datasets evaluated.

\myparagraph{Benchmarks.} Given the growing importance of transferability scorers, few works have focused on their evaluation.~\citet{agostinelli2022stable} proposed a large-scale benchmark for dataset and architecture transferability, with a resampling of experimental data reminiscent of ours, but introducing several \textit{ad-hoc} elements. The interpretation of their results is difficult, as usual notions of statistical analysis (e.g., standard deviations or confidence intervals) are not available for their sampling procedure. In contrast, we propose a streamlined benchmark design, with standard, interpretable statistical tools.

\section{Preliminaries}
\label{sec:preliminaries}

A complete choice of factors allowing to perform transfer learning is a \textbf{transfer scheme}. It may comprise factors such as datasets (source and target), tasks (source and target), model architecture, fine-tuning procedures, etc. Given our scope, our factors will be model architecture and target dataset.

A \textbf{transferability scorer} goal is to predict, without fine-tuning, the best scheme in a pool of candidates. Within our scope, the goal is to predict the best model architecture for a given target dataset.

The \textbf{benchmark} goal is to evaluate the scorers themselves, by measuring the correlation between their predicted transferability scores and the ground-truth \textbf{performance metric} of the architectures on the target datasets post fine-tuning. The measured correlation is the \textbf{benchmark outcome}. Our performance metric is  classification accuracy, and we discuss benchmark outcomes in Section~\ref{sec:benchmark_standard}.

\section{Back to Bayes: Combining Transfer Scores}
\label{sec:predictive}

Different scorers capture different theoretical and practical aspects of transferability (Section~\ref{sec:related}), but it is unclear how to combine them in a well-founded way. To compound that problem, scorers performance vary widely across datasets.

We propose a Bayesian hierarchical regression model to tackle those challenges, achieving a balance between the flexibility required to include all relevant information and the parsimony necessary to prevent overfitting. The model allows incorporating, in a principled way, knowledge about the behavior of scorers under known target datasets, in order to predict it for future unknown target datasets. Although Bayesian hierarchical models find a traditional and well-regarded usage in statistics~\cite{gelman2013bayesian, mcelreath2020statistical}, their use for transferability estimation is novel.

We assume that 
$(m,t,d,s)_i$, $1\leq{}i\leq{}N$ are calibration tuples with measured performance metric $m_i$ and transferability score $t_i$, both empirically z-normalized for each group of scorer~$\times$ dataset. The categorical variables $d_i$ and $s_i$ indicate target dataset and scorer, respectively. We denote as $\alpha_\bullet$ and $\beta_\bullet$ the intercept and slope of a linear regression, and as $\mu_\bullet$ and $\sigma_\bullet$ the mean and scale of a distribution. $\xi\sim\mathcal{D}(\theta)$ indicates $\xi$ is a random variable of distribution $\mathcal{D}$, parameterized by $\theta$, with $\p_\xi=\mathcal{D}(x;\theta)$ as the probability (density) of $\xi=x$. $\Norm$ is the Gaussian distribution parameterized by mean and standard deviation, and $\Exp$ is the exponential distribution parameterized by \textit{scale} (1/rate).

The model has three levels: one linear regression for each combination of scorer and dataset, the pooling of the parameters of all datasets for each scorer, and the pooling of all parameters.

\myparagraph{Likelihood.} The likelihood is a linear regression, with parameters per scorer~$\times$ dataset:
\begin{align}
    m_i &\sim \Norm(\mu_i, \sigma_{s_i,d_i}) &
    \mu_i &= \alpha_{s_i,d_i} + \beta_{s_i,d_i}\times{}t_i \label{eq:predictive_agnostic_ll}
\end{align}

\myparagraph{Level-1 priors:} for each combination of scorer $1\leq{}s\leq{}S$ and dataset $1\leq{}d\leq{}D$:
\begin{align}
    \alpha_{s,d} &\sim \mathcal{N}({\mu_\alpha}_s, {\sigma_\alpha}_s) &
    \beta_{s,d} &\sim \mathcal{N}({\mu_\beta}_s, {\sigma_\beta}_s) \nonumber &
    \sigma_{s,d} &\sim \Exp(\sigma_s)
\end{align}

\myparagraph{Level-2 priors:} for each scorer $1\leq{}s\leq{}S$:
\begin{align}
    {\mu_\alpha}_s &\sim \Norm(\mu_\alpha, \sigma_\alpha) &
    {\mu_\beta}_s &\sim \Norm(\mu_\beta, \sigma_\beta) \nonumber \\
    {\sigma_\alpha}_s &\sim \Exp(\sigma_\alpha) &
    {\sigma_\beta}_s &\sim \Exp(\sigma_\beta) &
    \sigma_s &\sim \Exp(\sigma)
\end{align}

\myparagraph{Level-3 priors:} pooling of all parameters:
\begin{align}
    \mu_\alpha &\sim \Norm(0,1) &
    \mu_\beta  &\sim \Norm(0,1) \nonumber \\
    \sigma_\alpha &\sim \Exp(1) &
    \sigma_\beta  &\sim \Exp(1) &
    \sigma &\sim \Exp(1)
\end{align}

\myparagraph{Posterior for parameters and predictions.} The posterior distribution does not have a simple analytic form. In Section~\ref{sec:predictive_details}, we discuss its numerical simulation.
Our goal is to estimate the distribution of the normalized performance metric $m$ given normalized transferability scores $(t,s)_j$, $1\leq{}j\leq{}S$ from the same scorers as before, all for the same (previously unseen) target dataset, and the same (potentially unseen) candidate architecture. This distribution has no analytic form either, but may be sampled from the same numerical simulation and ancestral sampling. 
         
The predicted normalized metric is then given by:
\begin{align}
    \p(\hat{m}) &= \frac{1}{S}\sum_j\p(\hat{m_j}|t_j,s_j) \label{eq:predictive_mixture} \\
    \p(\hat{m}_j|t_j,s_j) &= \int\Norm(\alpha + \beta{}t_j;\mu,\sigma)\times\mathrm{P} \label{eq:predictive_posterior}%
\end{align}
\noindent where the integral marginalizes all over the $\alpha$, $\beta$, $\mu$, and $\sigma$ parameters, their priors, and hyperpriors (omitted here and simply represented as $\mathrm{P}$, for clarity).

Remark that the resulting posterior has an arbitrary complexity, and, in particular, does \textit{not} assume the residuals are Gaussian. Through the level-2 priors, a scorer learns from multiple datasets, and through the level-3 priors, all scorers share knowledge. In the mixture of Eq.~\ref{eq:predictive_mixture}, the reliability of scorers is automatically weighted through the learned variability of their predictive distributions.

\subsection{Design and implementation details}
\label{sec:predictive_details}

Hierarchical models allow for expanded parameterization while avoiding overfitting. We can fit each combination of scorer and dataset with only 10 samples (the number of architectures). Through the hyperparameters, each estimation, too small in itself, shares enough knowledge with the others to be reliable.

The Gaussian and exponential distributions were chosen for practical and conceptual simplicity, as maximum entropy distributions and, thus, the least informative under suitable considerations. The choice of specific priors is not critical as long as they are neither too informative nor too lax. Although the \text{level-3} hyperparameters may seem narrow, consider that the variables in the regression are normalized, making large variations for the regression parameters extremely unlikely. We used prior and posterior predictive checks~\cite{mcelreath2020statistical} to ensure our choices were reasonable.

Computing the posterior parameter and posterior predictive distributions requires integrations that do not have an analytical form. We estimate them numerically, using Hamiltonian Monte Carlo (HMC)~\cite{neal1994improved}, implemented in the Stan probabilistic language~\cite{carpenter2017stan}. We used 4 chains, discarding 1000 samples as warm-up, and keeping 1000 samples for inference. The model samples very efficiently: in a modern multi-core computer, it takes less than 2 minutes to predict all 110 combinations of architectures and target datasets. We provide full implementation details in the appendix.

\myparagraph{Calibration data requirements.} Back to Bayes is relatively parsimonious in terms of its data requirements. In our experiments, the number of calibration tuples was equal to \textit{number of calibration scorers} $\times$ \textit{number of calibration datasets} $\times$ \textit{number of evaluated architectures}, resulting in, typically, 200-400 tuples, but sometimes as few as 30.

\section{Benchmark Design}
\label{sec:benchmarking}
We approach our benchmark as a statistical experimental design.
A transfer scheme (Section~\ref{sec:preliminaries}) plus a choice of transferability scorer is a treatment of the design, which we call a \textbf{benchmark treatment}. Since our aim is to benchmark the scorers themselves --- and not the transfer schemes ---  we consider the scorer as the factor of interest and the other factors (target dataset, model architecture) as nuisances.

\myparagraph{Benchmark outcome.}
\label{sec:benchmark_standard}
The outcome/response in our design is the agreement between the transferability scores and the performance metrics for each scorer, across architectures and target datasets.
Pearson's correlation coefficient, defined as $\pearson(\x,\y)=\cov(\x,\y)/\sigma_\x\sigma_\y$, is a straightforward measure of \textit{linear} correlation between two variables, but may grossly underestimate non-linear association.  The \textit{Kendall's tau} correlation accounts for any monotonic correlation by considering all possible pairs of the two variables, and counting the agreements and disagreements. Recent works (Table~\ref{tab:transferability-metrics} in appendix) have favored the \textit{weighted Kendall's tau}, penalizing more the disagreements on top-ranked elements. That reflects the importance of the top of the rank when choosing the transfer scheme. The weighted tau is defined as:

\begin{align}
  \tauw(\x,\y) &=\frac{ 
        \sum_{i<j}w(i,j)\sgn(x_i-x_j)\sgn(y_i-y_j)
        }{
        \sum_{i<j} w(i,j) 
        }  
        \label{eq:outcome_tauw} \\
    w(i,j) &= v(\rho_\x(x_i))+v(\rho_\x(x_j))+v(\rho_\y(y_i))+v(\rho_\y(y_j)) \\
    v(r) &= 1/(1+r)
\end{align}

\noindent where $\rho_\z(z_i)$ is the rank of the datum $z_i$ among the samples $\z$. The weighted tau reverts to the usual Kendall's tau by setting $w(i,j)=1$.

\myparagraph{Dealing with limited data}
Handling limited experimental data is a common challenge in the literature of transferability evaluation.
Indeed, reliably estimating the benchmark for a scorer requires a minimum of experimental samples, which translates to a modicum of datasets and architectures evaluated. However, each transfer treatment has to be carefully and individually fine-tuned, to provide a reliable ground-truth performance metric for the benchmark. We discuss below how to make the most of those expensive experimental data.

\myparagraph{Dataset as a nuisance.}
The target dataset is a statistical nuisance because the benchmark aims to evaluate the performance of the scorers not only for the exact datasets included in the study but also for future datasets similar to those. Therefore, a global assessment of the scorer across a representative selection of datasets is more important than the individual results on each dataset. 

Combining all datasets also pools more data, potentially improving the reliability of results. However, the tau statistics above become meaningless if one naïvely pools the data, because the comparison of pairs $(x,y)_i$ and $(x,y)_j$ is not pertinent if $i$ and $j$ point to different datasets (e.g., because some datasets are much easier than others). That difficulty has, so far, prevented combined analyses in the literature, where only per-dataset measurements are available.

We propose instead the \textbf{aggregated weighted tau}, which adapts the tau statistic to consider only the pairs within datasets. Considering a list of $N$ groups of data $\G=\{(\x_1,\y_1), \cdots, (\x_N,\y_N)\}$, where each pair $(\x_i,\y_i)$ is a complete set of samples $\{(x_j,y_j) \in (\x_i,\y_i) \}$. Then, the aggregated weighted tau is defined as:

\begin{align}
    \taug(\G) &=\frac{ 
         \sum_{(\x,\y)\in\G}\tau_\text{num}(\x,\y)
         }{
         \sum_{(\x,\y)\in\G}\tau_\text{den}(\x,\y)
        } \\
        \tau_\text{num}(\x,\y) &= \sum_{i<j}w(i,j)\sgn(x_i-x_j)\sgn(y_i-y_j)
        \\
        \tau_\text{den}(\x,\y) &= \sum_{i<j} w(i,j)
\end{align}

The combined analysis may be compared and contrasted with the \textit{averaged} analysis, in which the metrics are computed for each dataset, and \textit{then} averaged.
While averaging also allows marginalizing over the nuisance factor of datasets, it does not solve as well the issue of limited experimental data, as we will see in the experiments.

\myparagraph{Simulation.} 
Simulation allows making the most of limited experimental data, by preventing a peculiar data configuration or a single outlier from biasing the analysis. Bootstrapping, which we employ in this work, is the simplest form of simulation, where in each iteration, we sample (with replacement) a simulated data set of the same size as the original. The outcome may be measured across those iterations, providing a smoothed central measure (a mean or median) and an estimation of its spread (standard deviation, interquartile-range, or confidence interval). Beyond its simplicity, bootstrapping benefits from a long tradition of use in statistics, making it well understood. Yet, its application on the benchmarking of transferability, as proposed here, is novel.

\myparagraph{Source model raw metrics as a baseline.} Robust benchmarking require solid baselines. The findings of~\citet{kornblith2019better} suggest that the raw metrics of the source model should always be considered as a baseline, no matter the task. Those metrics are widely reported in transfer learning literature but are overlooked in the literature of transferability estimation up to this point. In our benchmark, we use the top-1 ImageNet accuracy as a baseline.

\section{Experiments}
\label{sec:experiments}

We applied the proposed benchmark (Section~\ref{sec:benchmarking}) to a selection of state-of-the-art transferability scorers and to the proposed Back to Bayes (Section~\ref{sec:predictive}). 

\myparagraph{SOTA scorers.} We selected all scorers applicable to architecture transferability that either had available source code, or whose experiments we could reproduce with our own code. If a scorer had multiple versions or free parameters, we selected the best according to author's original experiments. In total, we evaluated 13 SOTA scorers. Table~\ref{tab:transferability-metrics} in appendix summarizes the selection.

\myparagraph{Target datasets.} We evaluated the scorers for 11 target computer vision datasets, divided into four categories: generalist (Caltech-101~\cite{caltech101}, SUN397~\cite{sun397}, Pascal VOC 2007~\cite{voc2007}), fine-grained/natural (Oxford Flowers 102~\cite{flowers102}, Oxford-IIIT Pets~\cite{oxfordpets}), fine-grained/artifacts (FGVC-Aircraft~\cite{aircraft}, DTD~\cite{dtd}, Stanford Cars~\cite{stanfordcars}), and fine-grained/medical (BrainTumor~\cite{braintumor}, BreakHis~\cite{breakhis}, ISIC 19~\cite{isic19}). Those datasets cover a wide range of classes (2--397), training samples (2\,040--19\,850), and difficulty. 

\myparagraph{Target domain distance.} The four categories of our datasets represent four levels of domain distance to the source ImageNet dataset: generalist (the closest), fine-grained/natural, fine-grained/artifacts, and fine-grained/medical (the furthest), allowing us to apprehend the impact of domain shift on transferability. 

\myparagraph{Task.} Source and target tasks were always image classification.

\myparagraph{Architectures.} Our scorers chose from 10 architectures: ResNet-18,34,50~\cite{he2016deep}, DenseNet-121,161,169~\cite{huang2017densely}, MobileNetV2-0.5,1.0~\cite{sandler2018mobilenetv2}, EfficientNet-B0~\cite{tan2019efficientnet}, and ViT-small~\cite{dosovitskiy2020image}, covering 4~diverse architecture families, and important variants. We used implementations and checkpoints from Torchvision~\cite{torchvision2016}, except for the MobileNets and ViT-small, which were from TIMM~\cite{rw2019timm}.

\myparagraph{}The \textbf{benchmark outcome} is the weighted tau for the single-dataset and for the averaged experiments and the aggregated weighted tau for the combined experiments.

\myparagraph{Target fine-tuning.} Hyper-parameter search followed the Tuning Playbook~\cite{tuningplaybookgithub} guidelines, using Halton quasi-random sequences~\cite{halton_sequence}. Hyper-optimization was performed on a validation subset of the training set. The optimization always went for 100 epochs, with SGD with Nesterov momentum and cosine learning-rate scheduler.

We showcase the scenario of full-model fine-tuning, which is more used in practice, and more challenging for the scorers. We show results for the frozen-feature extractor scenario, as well as complete details on the training procedure, on the supplemental material.  

\subsection{SOTA evaluation}
\label{sec:experiments_sota}

To evaluate a scorer on a single dataset, we replicated the set of 10 measurements (one per architecture) over 1000 bootstrap iterations, computing the benchmark outcomes for each iteration. As our experiments considers 11 datasets, for the combined evaluation, we aggregated the 110 scores (10 measurements per dataset), replicated the entire set over 1000 bootstrap iterations, and computed the benchmark outcomes over each aggregated iteration. We used the traditional weighted and unweighted tau outcomes for the individual datasets and the averages, and the proposed aggregated tau for the combined evaluation.

\subsection{Back to Bayes evaluation}
\label{sec:experiments_btb}

Back to Bayes is intended as ``calibrate once/use many times'' and its calibration, once done, is intended to generalize to new scenarios. 
Back to Bayes is, thus, a principled way to combine existing scorers, by understanding them during its calibration phase. Here we evaluate those abilities.

\myparagraph{Choice of base scorers.} Our method is flexible enough to use all available scorer's information, or even as few as a single scorer. We evaluate Back to Bayes for a large choice of scorer combinations: each SOTA scorer together with the ImageNet predictor; the three top scorers (NCTI, ETran, PACTran) with and without ImageNet; the three bottom scorers  (SFDA, H-Score, R.H-Score) with and without ImageNet; and the 7 mid-range scorers in Figure~\ref{fig:ridgeline_wtau}. Although PACTran was not among the top-3 highest performers (by averaged or combined outcome), we decided to keep it on top due to its performance on the challenging medical datasets.

\myparagraph{Calibration.} We calibrated Back to Bayes generalization using a ``leave-one-dataset-out'' scheme to allow us to evaluate it on all target datasets. We also evaluate ablations with a single dataset for calibration (last three rows in Figure~\ref{fig:ridgeline_wtau}, calibration dataset indicated at the row header). We use the ``BtB + 3 Top'' configuration in the single-dataset experiments.

Again, Back to Bayes is relatively parsimonious in terms of its data requirements (Section~\ref{sec:predictive_details}). The experiments in Figure~\ref{fig:ridgeline_wtau} use 200 tuples for the SOTA experiments (rows 2--14), and then, for rows 15-22, respectively, 300, 400, 700, 300, 400, 30, 30, and 30 tuples.  

\myparagraph{Inference.} Back to Bayes' results are a probability distribution for the estimated standardized metric, represented as samples from its posterior obtained from the MCMC algorithm (Section~\ref{sec:predictive_details}). To output a single prediction, comparable with the other scorers, we averaged the samples, thus estimating the mean of the distribution.

\subsection{Findings}
\label{sec:results}

\begin{figure*}[ht!]
    \centering
    \includegraphics[width=\textwidth]{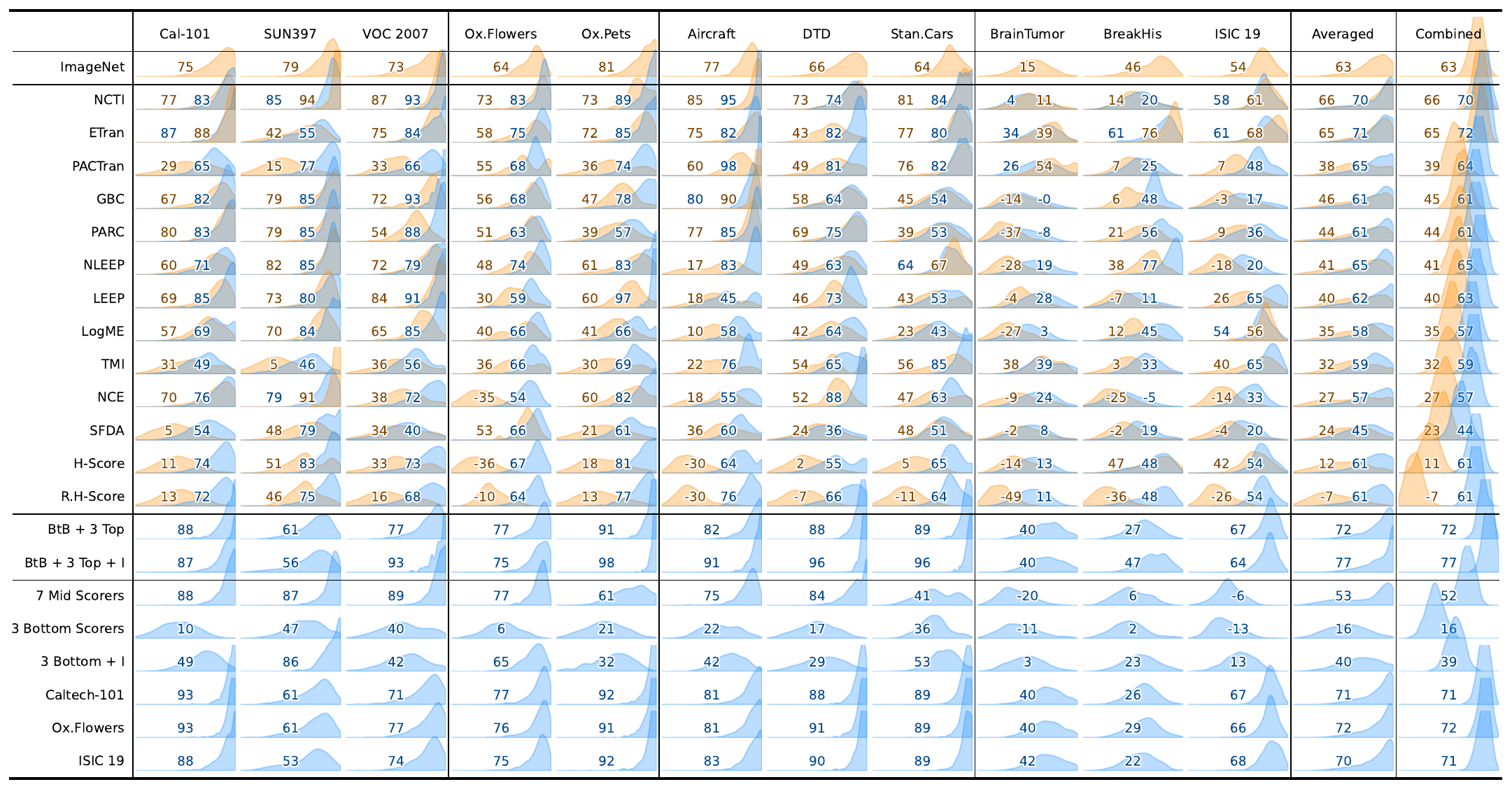}
    \caption{Weighted tau ($\tauw\times100$), higher is better. Row groups, from top: ImageNet baseline, state-of-the-art, Back to Bayes, ablations. Dataset groups, from left: generalist, natural, artifacts, medical. Averaged: average of individual dataset outcomes. Combined: outcome on combined dataset measurements, with aggregated weighted tau ($\taug\times100$). The ridgeline plots show the distribution of 1000 bootstrap iterations, whose mean appears in figures. Unmodified SOTA: orange; Back to Bayes: blue. (Best in color, details in the text.)}
    \label{fig:ridgeline_wtau}
\end{figure*}

Our results appear in Figures~\ref{fig:aggregated_plots}-\ref{fig:ridgeline_wtau}. To declutter the figures from zeros and decimal points, all benchmark outcomes appear $\times$100.

\myparagraph{Interpreting the results.} Figure~\ref{fig:ridgeline_wtau} shows, in each cell, the distribution of a thousand boostrapping samples for the weighted tau as well as the average value of those samples (in numbers). We use blue for the the proposed Back to Bayes, and orange for unmodified SOTA scorers. The isolated row on top is the ImageNet top-1 performance, used as a strong baseline.

The second row-group (NCTI to R.H-Score) shows the performance of SOTA scorers (orange) \textit{versus} the performance of Back to Bayes using only that single scorer combined with the ImageNet baseline (blue). 

The third row-group shows the best configurations of Back to Bayes: with the three ``top'' scorers (NCTI, ETRAN, and PacTRAN), and with those scorers and the ImageNet baseline.

The last row-group shows some ablations with different sets of scorers as well as Back to Bayes with the three ``top'' scorers, but calibrated on a single dataset (last three lines).

\textbf{Raw ImageNet metrics are a formidable baseline.} Contrasting the first row of Figure~\ref{fig:ridgeline_wtau} with the unmodified SOTA results (orange in the second row-group) shows that ImageNet presents better outcomes than most scorers. Only two scorers (NCTI and ETRAN) are competitive with that baseline, showcasing its importance. Both averaged and combined analysis columns show that relying on source performance will lead to good decisions in final transfer performance.    

\textbf{Averaged and combined (aggregated) taus have similar averages but different distributions.} The aggregated taus have much more peaked distributions over their bootstrapped samples, reflecting less uncertainty over the estimator. That showcases the advantage of the combined analysis, of pooling evidence from several experiments, on providing more statistically reliable estimations for the tau.

\textbf{Back to Bayes and ImageNet improves scorers consistently.} As the first group of rows below ImageNet in Figure~\ref{fig:ridgeline_wtau} shows, single scorers benefit consistently of incorporating knowledge of ImageNet using Back to Bayes. NCTI, which was the best performing unmodified scorer, presented significant improvements when combined with ImageNet through BtB, with both averaged and combined $\tauw$ going from 66 to 70. Poor performing scorers showed even more significant improvements, becoming comparable with the best ummodified best ones after using BtB. H-Score's combined $\tauw$, for example, went from 11 to 61. 

\textbf{Back to Bayes+the 3 top scorers provided the most dependable results}, with narrower variation around a desirable mean. That was true whether or not ImageNet (``+I'' on the table) was added to the mix.

\textbf{Back to Bayes can learn from little data and is strongly regularized}. As shown in the ablations of the last three lines of Figure~\ref{fig:ridgeline_wtau}, calibrated on a single dataset, with as few as 30 calibration tuples. The ``Caltech-101'' row for example, uses only the performances and scorers for this dataset, achieved by the Top 3 Scorers over the 10 analyzed architectures. Still, the combined and averaged $\tauw$ reached 71 points, being well above any previous unmodified scorer result, which failed to surpass the 66 points of NCTI.

\textbf{Back to Bayes is not garbage-in-gold-out}, as demonstrated by the ablations with the bottom 3 scorers. The distributions in Figure~\ref{fig:ridgeline_wtau} for the ``3 Bottom Scorers'' show much more flattened curves and lower peaks compared to Mid or Top Scorers, or to when ImageNet samples are added (``3 Bottom + I''). The information exploited has to come from \textit{somewhere}: either from the quality of the scorers or their (learnable) complementarity. 

\textbf{Medical datasets are still challenging for everyone.} The three medical datasets are generally where all predictors — including ImageNet — have worse and most inconsistent performance, with ETran, arguably, being the lone exception.

\section{Conclusions and Future Directions}
\label{sec:conclusions}

The increasing relevance of transfer learning is boosting interest in transferability estimation, accompanied by techniques advancing in both theoretical sophistication and practical performance. Our work adds to this conversation by proposing a well-founded method to merge scorers and a streamlined benchmark for scorers.

The continued difficulties of scorers on medical datasets suggest that other specific applications may also be blind spots when benchmarking transferability scorers, and that a diverse panel of specialist datasets might be critical to evaluate their generalization abilities.

\myparagraph{Limitations.}
Currently, Back to Bayes's rich uncertainty information output is underutilized, as we simply employ the mean from the MCMC samples. 
Fully exploring this uncertainty in downstream tasks is a promising avenue for future research. 

Back to Bayes requires, for calibration, the ground-truth performance metric for the scorers over a panel of architectures and datasets. Remark, however, that once calibrated, Back to Bayes no longer requires this information: it is meant to be ``calibrate once, use many times''. Back to Bayes can generalize from as few as 30 calibration tuples. 

We purposefully restricted our scope to architecture transferability, an essential scenario that intersects most of the current literature and has many practical applications. Since most of the state-of-the-art transfer scorers fall short of a simple ImageNet baseline even within that restricted boundary, adding factors to our research, although tempting, appeared premature at this point. Nevertheless, as  advanced scorers (such as NCTI and ETRAN) appear, and as techniques based on learning (such as Back to Bayes) start to overperform raw ImageNet scorers, future works might want to address more complex scenarios.

\myparagraph{Acknowledgments:}  L. Chaves is funded by Becas Santander/UNICAMP – HUB 2022, Google LARA 2021, in part by the Coordenação de Aperfeiçoamento de Pessoal de Nível Superior – Brasil (CAPES) – Finance Code 001. A.~Bissoto is funded by FAPESP (2019/19619-7, 2022/09606-8). S.~Avila is partially funded by CNPq PQ-2 grant 316489/2023-9, FAPESP 2013/08293-7, 2020/09838-0, 2023/12086-9, and H.IAAC (Hub de Inteligência Artificial e Arquiteturas Cognitivas), and Google AIR~2022.

\bibliographystyle{abbrvnat}
\bibliography{main}

\appendix
\newpage
\section{Appendix / Supplemental Material}
\renewcommand{\thesection}{A\arabic{section}}
\setcounter{section}{0}

\section{Frozen Feature Extractor} 

\begin{figure*}[ht!]
    \centering
    \includegraphics[width=\textwidth]{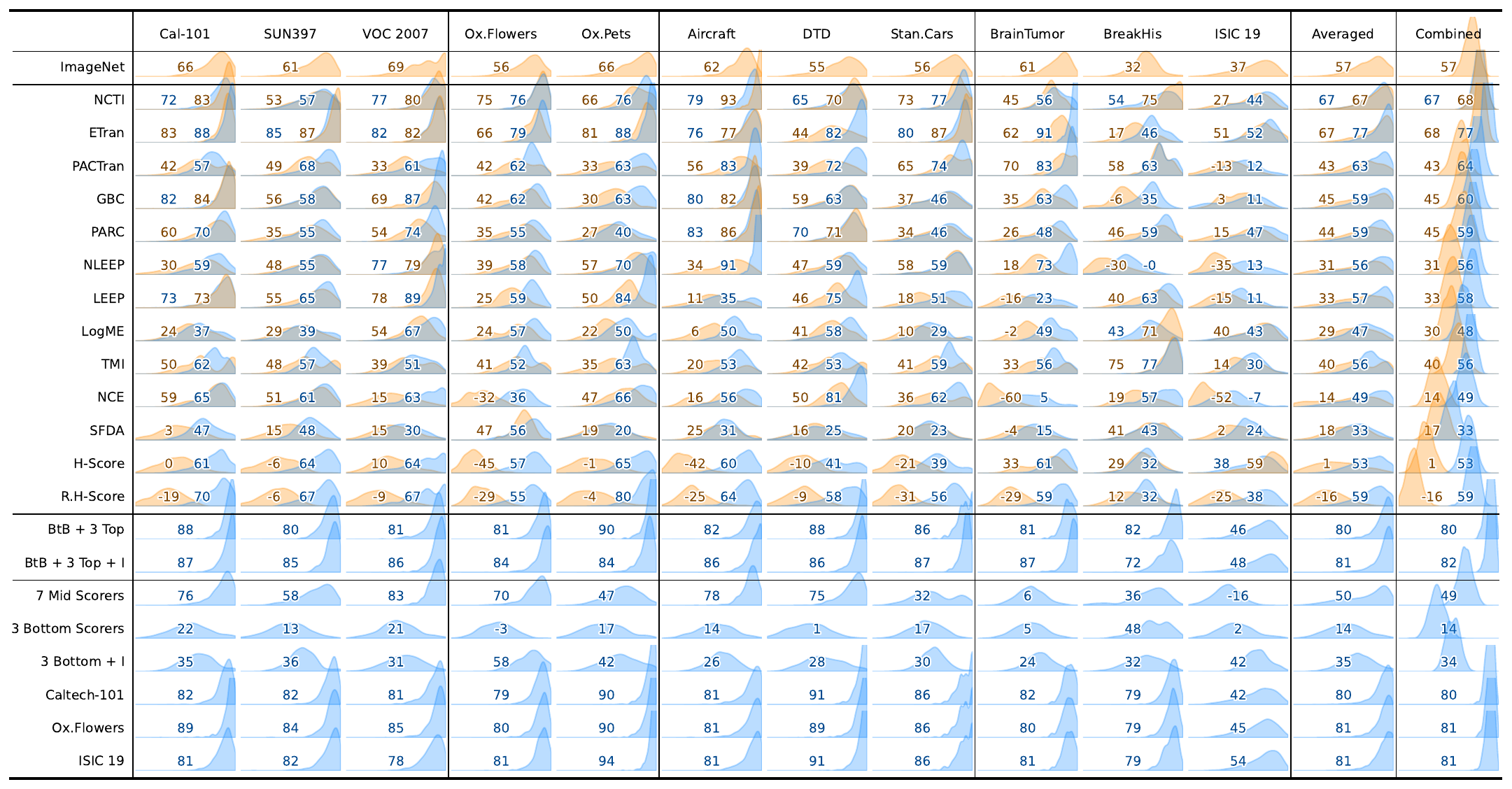}
    \caption{Frozen feature extractors, weighted tau ($\tauw\times100$), higher is better. Row groups, dataset groups, and colors as in Figure~\ref{fig:ridgeline_wtau}).}
    \label{fig:frozen_ridgeline_wtau}
\end{figure*}

In the main text, we focus on the scenario of full-model fine-tuning, which is most often used in practice for the relatively small models used in computer vision. For the sake of exhaustiveness, we provide the results for a frozen feature extractor and learned last-layer-classifier in this appendix. The results appear in Figure~\ref{fig:frozen_ridgeline_wtau}. Theoretically, this should be an easier scenario for most scorers, but we do not observe a systematic improvement: fluctuations vary almost randomly per scorer and dataset. However, the global trends and findings are still the same as those of the main experiment (Section~\ref{sec:experiments}).

\section{Back to Bayes Implementation Details}
\label{sec:supp:btb_implementation}

We implemented Back to Bayes using the Python bindings PyStan v3.9 for Stan v2.32 on Python~3.10. We used ArviZ v0.17 for summarizing and diagnosing the chains, using well-known diagnostic statistics such as effective sample sizes and Gelman-Rubin's R-hat~\cite{gelman2013bayesian}, as well as diagnostic plots for the chains. We also employed prior and posterior predictive checks to ensure the model was well fit~\cite{gelman2013bayesian,mcelreath2020statistical}. Because hierarchical models with a naive parameterization sample inefficiently, we adapted our models to use non-centered parameterization. The procedure is akin to the ``reparameterization trick'' commonly applied to probabilistic deep models. We sampled from 4 chains, discarding 1000 samples as warm-up and keeping 1000 samples for inference, for a total of 4000 samples per inference. 

Although we separate, conceptually, the \textit{calibration} and \textit{inference} of the model (Section~\ref{sec:experiments_btb}), in practice, because the posterior of the model parameters has no analytic form, those steps must happen at once: we ``fit'' the posterior with the training data at the same time we sample from the posterior of the predictive distribution for the new test data. Each of the 4000 MCMC samples gives, at once, a sample from the posterior for each parameter and a sample from the posterior of each test datum. The reparameterized model samples efficiently and each chain may be sampled independently. 

Thanks to parallelism, it takes less than 2 minutes to predict all 110 combinations of architectures and target datasets in a server with 40 Intel Xeon CPUs at 2.20GHz.

\section{Deep Models Fine-Tuning Details}
\label{sec:supp:fine-tune-details}

We adopted the same protocol for training the transfer treatments that is adopted throughout the transfer scoring literature~\cite{logme, sfda,pactran,etran,ncti,tmi}.The transfer treatments consisted of varying choices of model architecture and target dataset. The source dataset was always  ImageNet-1K (ViT-small is pre-trained on ImageNet-21K and fine-tuned on ImageNet-1K), and the source and target tasks were always image classification.

\paragraph{Hyperparameter tuning:} We followed the Tuning Playbook~\cite{tuningplaybookgithub} guidelines, using quasi-random sampling (in our case, Halton sequences~\cite{halton_sequence}) for the hyperparameters. We used SGD as the optimizer, cosine learning-rate scheduler, 100 epochs, and batch size of 128. We searched over 75 quasi-random combinations of learning rate in the range [$10^{-4}, 10^{-1}$] and weight decay in the range [$10^{-6}, 10^{-4}$] for each model architecture and dataset.

When the datasets provided official train/validation splits, we used them for the hyperparameter search. Otherwise, for generalist datasets, we used a random 80:20 train/validation split. For medical datasets, we used a stratified 80:20 train/validation. 

\paragraph{Architecture training:} Training data augmentations comprised random horizontal and vertical flips, random resized crops to a target size of $224\times{}224$ using scaling in $[0.75, 1.00]$, random rotations of up to 45 degrees, and hue color jitter of up to 20\%. For validation, test, and scorer evaluation, we resized the input images to $256\times{}256$, then took a center crop of $224\times{}224$.

\paragraph{Computing:} We run the experiments on NVIDIA RTX 5000 and RTX 8000 GPUs. We select the best-performing model in the validation set for each architecture for test evaluation. In total, we trained $16\,500$ models, including the hyperparameter search. 

\paragraph{Data License:} We use only publicity available datasets. BreakHis~\cite{breakhis}, BrainTumor-Cheng~\cite{braintumor}, Caltech-101~\cite{caltech101} are under CC BY 4.0 license, while ISIC2019~\cite{isic19} is under CC BY-NC 4.0, and Oxford-IIIT Pets~\cite{oxfordpets} is under CC BY-SA 4.0. We were unable to find the license for remaining datasets. 

\section{Source Code}

We will link our code repository in the final version. 
The code comprises the deep-models training, scorers computation, and Back to Bayes implementation.

\begin{table*}[ht]
\scriptsize
\caption{Summary of transferability scoring methods. Bench: benchmark outcomes (Section~\ref{sec:benchmark_standard}).  ST: source model training: Supervised or Self-\textit{s}up. T: task: Classification, Object-detection, Regression, Segmentation. RT: approximate runtime (s) for full run of 110 schemes.} 
\label{tab:transferability-metrics}
\centering
\begin{tabular}{
>{\raggedleft\arraybackslash}p{1.60cm}
>{\raggedright\arraybackslash}p{0.4cm}
>{\raggedright\arraybackslash}p{5.2cm}
>{\raggedright\arraybackslash}p{1.0cm}
>{\raggedright\arraybackslash}p{0.5cm}
>{\raggedright\arraybackslash}p{0.5cm}
>{\raggedright\arraybackslash}p{0.75cm}
>{\raggedleft\arraybackslash}p{0.35cm}}
\midrule
 & Year & Details & Metric & Bench &ST&T&RT \\ \midrule 

H-Score~\cite{hscore} & 2019 & Transferability correlates to inter-class variance and feature redundancy & Acc & $\rho$ & S & C & 30\\ 

NCE~\cite{nce} & 2019 & Negative conditional entropy between source and target labels & Acc & $R_{p}$ & S & C & $<$2  \\ 

LEEP~\cite{leep} & 2020 &Log-likelihood between target labels and the idealized classifier & Acc & $R_{p}$ & S & C & $<$2 \\ 

$\mathcal{N}$LEEP~\cite{nleep} & 2020 & Log-likelihood between target labels and Gaussian mixture from target's dataset features & Recall$@k$, Rel@$k$ & $\tauv$, $R_{p}$  & S, Ss & C, O, S & 600 \\ 

PARC~\cite{parc}& 2021 & Computes the Pearson product-moment correlation between the features and the labels & Acc & $R_{p}$ & S & C, O & 100 \\

R.~H-Score~\cite{reghscore} & 2021 & Improves H-Score by applying shrinkage estimators for stable covariance & Acc & $R_{p}$ & S & C &40 \\ 

SFDA~\cite{sfda}  & 2021& Leverages Fisher Discriminant analysis to improve class separability and self-challenging mechanism with hard-examples  & Acc & $\tauw$  & S, Ss & C & 40 \\

LogME~\cite{logme} & 2021& Learns a Bayesian linear classification and calculates the maximum evidence & Acc & $\tauw$   & S, Ss & C, R &30 \\ 

GBC~\cite{gbc} & 2021 & Computes the features overlap for each class using Bhattacharyya coefficient & Acc & $\tauw$, $\tauv$, $R_{p}$ & S & C, S &10 \\ 

PACTran~\cite{pactran} & 2022 & Seeks an optimal yet efficient PAC- Bayesian bound to the generalization error based on the cross-entropy   & Acc & $\tauv$ & S & C &200 \\

ETran~\cite{etran}  & 2023 & Hybrid scorer based on energy and Linear Discriminant Analysis to determine whether the target dataset is OOD or IID & Acc & $\tauw$ & S & C, O &40 \\

NCTI~\cite{ncti} & 2023 & Measures how close the features are from the neural collapse state & Acc & $\tauw$, $\rho$ & S, Ss & C &20 \\

TMI~\cite{tmi} & 2023 & Measures intra-class feature variance using conditional entropy & Acc &  $\tauv$ & S, Ss & C & 20 \\ 
\midrule
\end{tabular}

\end{table*}

\end{document}